\documentclass[11pt,a4paper]{article}
\usepackage[hyperref]{emnlp-ijcnlp-2019}
\usepackage{times}
\usepackage{latexsym}
\usepackage{todonotes}

\usepackage{url}
\usepackage{booktabs}

\usepackage{graphicx}
\usepackage{subcaption}
\aclfinalcopy % Uncomment this line for the final submission

\title{Fact-Checking Meets Fauxtography: Verifying Claims About Images}

\author{Dimitrina Zlatkova \\
  Sofia University \\ 
  ``St. Kliment Ohridski'' \\
  Sofia, Bulgaria \\
  {\tt dvzlatkova@uni-sofia.bg} \\\And
  Preslav Nakov \\
  Qatar Computing Research Institute, \\
  HBKU \\
  Doha, Qatar \\
  {\tt pnakov@qf.org.qa} \\\And
  Ivan Koychev \\
  Sofia University \\ 
  ``St. Kliment Ohridski'' \\
  Sofia, Bulgaria \\
  {\tt koychev@fmi.uni-sofia.bg} \\
  }

\date{}

\begin{document}
\maketitle
\begin{abstract}
The recent explosion of false claims in social media and on the Web in general has given rise to a lot of manual fact-checking initiatives.
Unfortunately, the number of claims that need to be fact-checked is several orders of magnitude larger than what humans can handle manually. Thus, there has been a lot of research aiming at automating the process. 
Interestingly, previous work has largely ignored the growing number of claims about images. This is despite the fact that visual imagery is more influential than text and naturally appears alongside fake news.
Here we aim at bridging this gap. In particular, we create a new dataset for this problem, and we explore a variety of features modeling the claim, the image, and the relationship between the claim and the image. The evaluation results show sizable improvements over the baseline. We release our dataset, hoping to enable further research on fact-checking claims about images.
\end{abstract}

\section{Introduction}

As social media become a bigger part of our daily lives, their influence over the way people think and make decisions increases. Inevitably, this has offered opportunities for fake content to arise and to spread faster than ever, e.g., recent research has shown that fake news spreads six time faster than real news \cite{Vosoughi1146}. Sometimes such content is created for pure entertainment or for financial gain from advertisement shown alongside the fake content, but more often and especially recently it has been used to spread disinformation, e.g.,~with the aim to influence political elections \cite{CoNLL2019:troll:roles}.
To deal with the problem, a number of manual fact-checking initiatives have been launched, but they remain insufficient to cope with the ever growing number of check-worthy claims. Thus, automated methods have been proposed as a more scalable solution.

\begin{figure}[tbh]
\begin{subfigure}{.25\textwidth}
  \centering
  \captionsetup{width=.9\linewidth}
  \includegraphics[width=.95\linewidth]{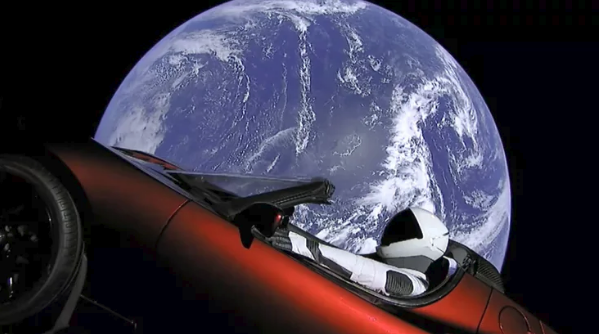}
  \caption{A series of images show a Tesla vehicle in space.}
   \label{fig:true_true_a}
\end{subfigure}%
\begin{subfigure}{.25\textwidth}
  \centering
  \captionsetup{width=.9\linewidth}
  \includegraphics[width=.9\linewidth]{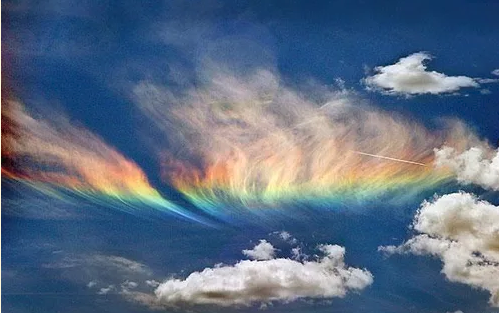}
  \caption{Photograph shows a “fire rainbow” over Idaho.}
\end{subfigure}
\caption{Examples of Real images and True claims.}
\label{fig:true_true_examples}
\end{figure}

Recently, a growing number of claims have been about images.
The word \textit{Fauxtography} has been used to describe images, especially news photographs, that convey a questionable, or outright false, sense of the events they seem to depict. The term was coined over a decade ago \cite{DCooper}, and there is growing research interest in the topic in the Computer Vision community \cite{Bayar2016ADL, Carvalho2016IlluminantBasedTS}.
Given the recent proliferation of fake news, and given that many of the questionable claims are about images, it would be natural to expect similar interest in the Computational Linguistics community, especially given the fact that visual imagery is more influential than text and naturally appears alongside fake news. Yet, computational fact-checking has mostly ignored the growing number of claims about images. 
Here we aim at bridging this gap. In particular, we create a new dataset for this problem, and we explore a variety of features modeling the claim, the image, and the relationship between the claim and the image.

Let us look at some examples.
Figure~\ref{fig:true_true_examples} shows two images that look surrealistic, and thus spark interest and raise natural suspicion. Yet, they are in fact real.\footnote{\url{http://www.snopes.com/fact-check/tesla-car-really-space}}$^{,}$\footnote{\url{http://www.snopes.com/fact-check/fire-rainbow}}

\begin{figure}[tbh]
\begin{subfigure}{.25\textwidth}
  \centering
  \captionsetup{width=.9\linewidth}
  \includegraphics[width=.9\linewidth]{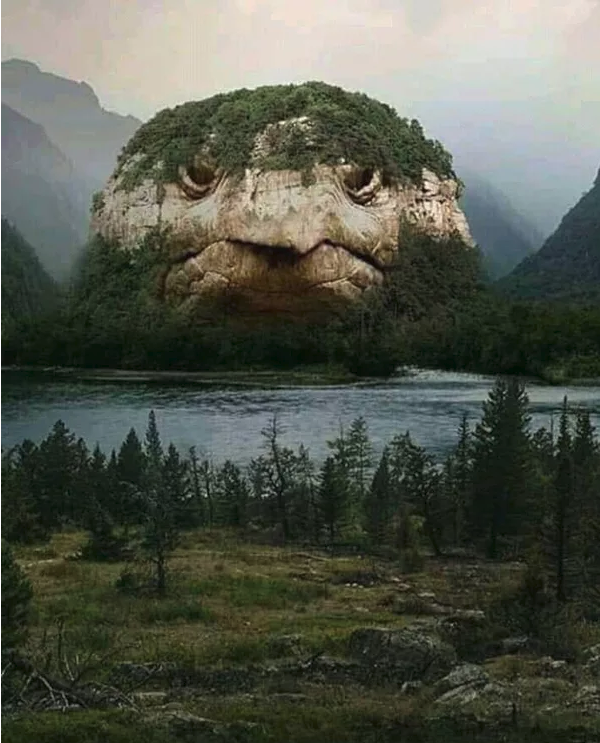}
  \caption{A photograph shows a mountain resembling a turtle.}
  \label{fig:false_false_a}
\end{subfigure}%
\begin{subfigure}{.25\textwidth}
  \centering
  \captionsetup{width=.9\linewidth}
  \includegraphics[width=.9\linewidth]{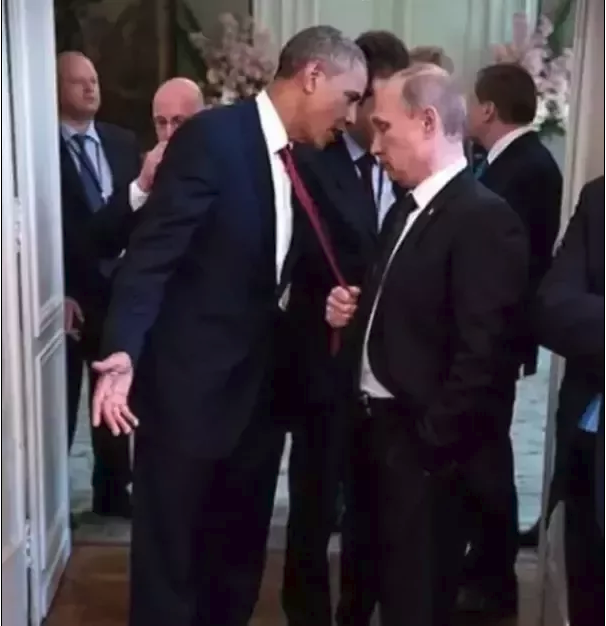}
  \caption{A photograph shows Russian president Vladimir Putin aggressively pulling on U.S. president Barack Obama's tie.}
  \label{fig:false_false_b}
\end{subfigure}
\caption{Examples of Fake images and False claims.}
\label{fig:false_false_examples}
\end{figure}

Figure~\ref{fig:false_false_examples} contains fake images accompanied by false claims. On the left, in Figure~\ref{fig:false_false_a}, we see an image from a Facebook page that claims to show a turtle mountain. It turns out that this is not a genuine photograph of a real-world location, but a digital artwork comprising altered versions of at least two different photographs.\footnote{\url{http://www.snopes.com/fact-check/turtle-mountain-photo/}} On the right, Figure~\ref{fig:false_false_b} displays an image purportedly showing Putin aggressively grabbing President Obama by the tie and pulling him close. This image has been digitally manipulated.\footnote{\url{http://www.snopes.com/fact-check/putin-obama-tie-pull/}}

\begin{figure}[tbh]
\begin{subfigure}{.28\textwidth}
  \centering
  \captionsetup{width=.9\linewidth}
  \includegraphics[width=.9\linewidth]{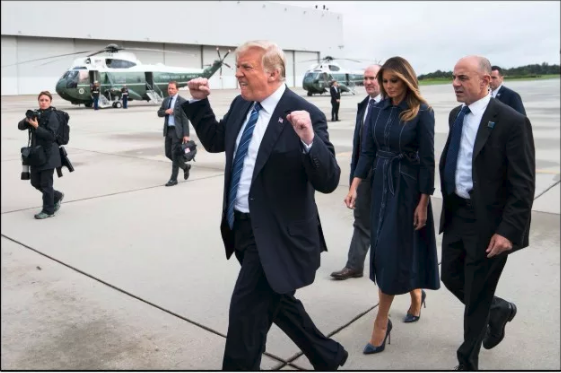}
  \caption{President Trump's notorious `fist pump' at a Pennsylvania airport on 9/11 was offered as a greeting to a cancer victim.}
  \label{fig:false_true_a}
\end{subfigure}%
\begin{subfigure}{.22\textwidth}
  \centering
  \captionsetup{width=.9\linewidth}
  \includegraphics[width=.9\linewidth]{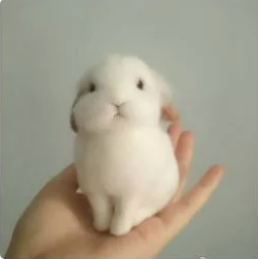}
  \caption{A photograph shows a palm-sized rabbit.}
  \label{fig:false_true_b}
\end{subfigure}
\caption{Examples of True images and False claims.}
\label{fig:false_true_examples}
\end{figure}

Finally, Figure~\ref{fig:false_true_examples} shows original photographs with false claims about them. The image in Figure~\ref{fig:false_true_a} shows Trump with his fists in the air, but his gesture is not a greeting to a cancer victim as the claim states. The real image was used as a part of a meme\footnote{\url{http://www.snopes.com/fact-check/trump-fistpump-cancer-greeting/}} that was designed to make it seem that way. The photo in Figure~\ref{fig:false_true_a} was posted by a Twitter account in an attempt to go viral, claiming that it shows a real bunny sitting in the palm of someone's hand. It actually shows a plush doll.\footnote{\url{http://www.snopes.com/fact-check/bunny-toy-photograph/}}

As we have seen above, there are a number of reasons why an image may be deemed fake. In most cases, this involves some kind of digital manipulation, e.g.,~cropping, splicing, etc. However, there are cases when an image is completely legitimate, but it is published alongside some text that does not reflect its content accurately. This is our main focus here: we study the factuality of image-claim pairs.

The contributions of this paper can be summarized as follows:
\begin{itemize}
    \item We study a new problem: predict the factuality of a claim with respect to an image.
    \item We create a new dataset for this problem, which we release to the research community in order to enable further work.
    \item We explore a variety of features, and we demonstrate sizable improvements over the baseline.
\end{itemize}

The remainder of this paper is organized as follows: Section~\ref{sec:related} presents some relevant related work. Section~\ref{sec:method} describes in depth our method and the various features we experimented with. Section~\ref{sec:data} gives details about the datasets we created and used. Section~\ref{sec:experiments} describes our experimental setup and presents the evaluation results. Section~\ref{sec:discuss} gives additional details about the performance of the individual features, both in isolation and in various combinations, and further describes some unsuccessful attempts at extracting better features. Finally, Section~\ref{sec:future} presents our conclusions and some ideas for future work.

\section{Related Work}
\label{sec:related}

\subsection{Fact-checking Claims}
There has been a lot of research in the last few years in automatic fact-checking of claims and rumors, which can be classified into two general categories. The first approach focuses on the social aspects of the claim and how users in social media react to it \cite{Canini:2011,Castillo:2011:ICT:1963405.1963500,ma2016detecting,PlosONE:2016,P17-1066,dungs-EtAl:2018:C18-1}. This is reflected by user comments, likes/dislikes, views and other types of reactions, which are collected and used as features. 

\noindent Other methods use the Web and try to find information that proves or disproves the claim \cite{mukherjee2015leveraging,Popat:2017:TLE:3041021.3055133,RANLP2017:factchecking:external,AAAI2018:factchecking,baly-EtAl:2018:N18-2}. In either case, what is important is the stance \cite{riedel2017simple,thorne-EtAl:2017:NLPmJ,hanselowski-EtAl:2018:C18-1,NAACL2018:stance,EMNLP2019:Stance:crosslanguage:contrastive}: whether the opinion expressed in a tweet or in an article by a particular user/source agrees/disagrees with the claim, and the reliability of the source, i.e.,~can we trust this source \cite{D18-1389,source:multitask:NAACL:2019}. 

We should note that all these approaches are limited to textual claims, while we are interested in claims about images.

\subsection{Detecting Fake/Manipulated Images}

The task of detecting fabricated images falls under the area of image forensics. Such tasks are usually solved using traditional statistical methods modeling color, shape, and texture features \cite{bayram2006image,stamm2010forensic,Carvalho2016IlluminantBasedTS}. More recently, with the rise of Deep Learning, modern approaches and architectures have been applied to tackle the problem \cite{Bayar2016ADL}. However, most existing work uses datasets with generic images and very few papers specialize in the area of news and social media \cite{Zhiwei2016}. Detecting manipulation in the images is relevant for us, but is not enough, since often the image is original, but the claim about it is false.

\subsection{Fact-Checking Claims about Images}
Little research exists on the topic of fact-checking claims about images, where the input to be analyzed is an image-claim pair. To the best of our knowledge, there is only one work closely related to ours: a recent paper \cite{zhang2018fauxbuster} presents a system called FauxBuster which aims to fight against Fauxtography. We differ from them in that we use the Web as a source of information. In contrast, they focus on the social aspects of the problem and use comments on Twitter and Reddit to extract features, which makes our work complementary to theirs. Unfortunately, direct comparison to their approach is not feasible, as their dataset is not freely accessible.

\section{Method}
\label{sec:method}

This section describes the different approaches we applied towards engineering and extracting features from the image-claim pair.

We start with \emph{reverse image search}. The classical image search allows users to search for images based on a text with specific words or phrases. In contrast, reverse image search takes as input an image and returns Web pages that include this exact image or images that are very similar to it. This process can be easily automated and applied to a large number of images via Google's Vision API.\footnote{\url{http://cloud.google.com/vision/}} It can also return other information related to the image, e.g.,~tags, the text on the image, some object detection, explicit content, etc.

Using reverse image search, for each image we obtain a maximum of 50 Web pages that contain it. We remove pages that are known to be from fact-checking Web sites such as snopes.com, factcheck.org, using open-source code.\footnote{\url{github.com/clef2018-factchecking}} For the remaining Web pages, we crawl the article and we get its title and text.

\subsection{Features about the Image}

{\bf Google tags}: This is a list of tags that Google associates with the image. We decided to use this list because it contains words and phrases about events and people related to the image, which might give us an insight about what the image contains and what it is about. For example, the image in Figure~\ref{fig:true_true_a} has the following tags: \emph{SpaceX, Falcon Heavy, Rocket, Rocket launch, Falcon, Company, Launch pad, Booster, Thrust, Entrepreneur, Elon Musk}. After lowercasing them and removing stop words, we use them directly as bag-of-words features.

{\bf URL domains}: The Web pages that contain the image usually come from media sources and represent articles on a topic related to the image and/or the claim attached to it. However, in some cases they might point to an image-hosting service or a social network Web site such as Pinterest, Imgur, Twitter, etc. In an effort to use this fact, we extracted the top-level domain names from the list of URLs and we used them as TF.IDF features.

{\bf URL categories}: In order to get more insight about what types of websites write about fake and genuine images, we classify them in several predefined URL categories.

\noindent We use open-source code\footnote{\url{http://github.com/matthewruttley/mozclassify}} to classify URLs, which performs rule-based matching of tokens from the URL against a predefined list of words. Given a URL, it assigns it a tuple of one higher-level and one lower-level category. For example, when we run the algorithm on the Web sites returned for the image in Figure~\ref{fig:true_true_a} we get category tuples such as: (`arts \& entertainment', `general'), (`sports', `general'), (`society', `general'), (`technology \& computing', `general'), (`science', `general'), (`automotive', `general') and (`business', `marketing'). To transform those into features, we take all Web sites returned by the reverse image search for the image, and we merge the lists of their category tuples. We do not differentiate between high- and low-level categories; rather, we just apply TF.IDF on the combined list.

{\bf True/False/Mixed media percentage}: In order to determine whether an image is fake or not, we can also check the reliability of the sources that wrote about it. \emph{Media Bias/Fact Check}\footnote{\url{http://mediabiasfactcheck.com/}} (MBFC) is a Web site that provides factuality information about 2700+ media sources. We use their database to classify each Web page that is returned by the reverse image search into the following categories: \emph{True} (high factuality), \emph{False} (low factuality) and \emph{Mixed} (mixed factuality). Then, we use the percentage of Web pages from each category returned by the reverse image search as a feature.

{\bf Known media percentage}: If a URL is not on the MBFC list, we label it as \textit{Unknown} and we use the percentage of known Web pages as a feature.

{\bf True/False/Mixed media titles}: We use the titles of the articles from a True, False or Mixed media as bag-of-words features.

\subsection{Features about the Claim}

So far, in our feature extraction process we have only used the image from the image-claim pair, which means we might be missing crucial information. After manual inspection of a few examples, we realized that about half of them can be classified only using the image, e.g.,~because it is a collage, was photoshopped, or manipulated in some way. The other half contain legitimate images that might appear on trustworthy Web sites, but the claim associated with them was false.

{\bf Claim text}: We transform the text of the claim into a TF.IDF vector, which we use as a feature.

\subsection{Features about the Image-Claim pair}

In addition to using the claim text, we want to check how it is related to the image and whether the claim is true with respect to it. We model that by comparing the text of the claim to the articles returned by the Reverse Image Search of the image. We use only the articles from trustworthy media sources, according to our MBFC labels. We approach the task of computing the similarity of those texts in two different ways.

{\bf Cosine similarity}: We perform the comparison on the TF.IDF representations of the claim and each article's title. We compute a smoothed average on the list of cosine similarities to get the final feature value.

% If $X$ is the vector of the claim and $Y$ is a list of vectors for each title, the cosine similarity is computed as the normalized dot product of $X$ and $Y$.\todo{I am confused here: $X$ is a vector, but $Y$ is a list of vectors? Cosine similarity is well-known. Can't we just say that we have average cosine similarity?} This gives us a list of similarities $S$ on which we apply a smoothed average to get the final feature value using the formula:\todo{This is a strange smoothing: you only increase the denominator, not the nominatior.}

% $$\frac{\sum S}{1 + \left | S \right |}$$\todo{As the denominator is confusing, maybe we should skip this formula and just say we use smoothed average without the formula.}

{\bf Embedding similarity}:
We use pretrained embeddings of size 512 \cite{cer-etal-2018-universal} as a way to vectorize the claim and the title sentences. Then, it is trivial to calculate the similarity as a dot product, as they are already in a normalized form. Again, we use a smoothed average to reduce the list of similarities to a single number.

\section{Data}
\label{sec:data}

As we have a new task, we needed to create our own dataset. In fact, we created two datasets from two separate sources, but with similar qualities and format.  The main idea behind the data collection process was to find viral, interesting and even contradictory images with some text that describes them, i.e.,~the \textit{claim}. Both datasets are in English.

\subsection{The Snopes Dataset}

Snopes.com is arguably the oldest and the largest fact-checking Web site online. It aims to fight misinformation by investigating different pieces of news. The site has a special section for image-related fact-checking, called \textit{Fauxtography}\footnote{\url{http://www.snopes.com/fact-check/category/photos}}. It uses an extensive list of labels to classify each piece of news as \emph{True}, \emph{False}, \emph{Miscaptioned}, \emph{Mixture}, \emph{Undetermined}, \emph{Unproven}, \emph{Outdated}, etc. For the purpose of our dataset, we gather only image-claim pairs that were labeled as either \emph{True} or \emph{False}. The collected data consists of \textbf{838 examples} of which 197 \emph{True} and 641 \emph{False}. The huge imbalance of the classes might be surprising at first, but it makes sense for fact-checkers to prefer to spend their time fact-checking news pieces that have a higher chance of being fake.

\begin{figure}[tbh]
\begin{subfigure}{.25\textwidth}
  \centering
  \captionsetup{width=.9\linewidth}
  \includegraphics[width=.9\linewidth]{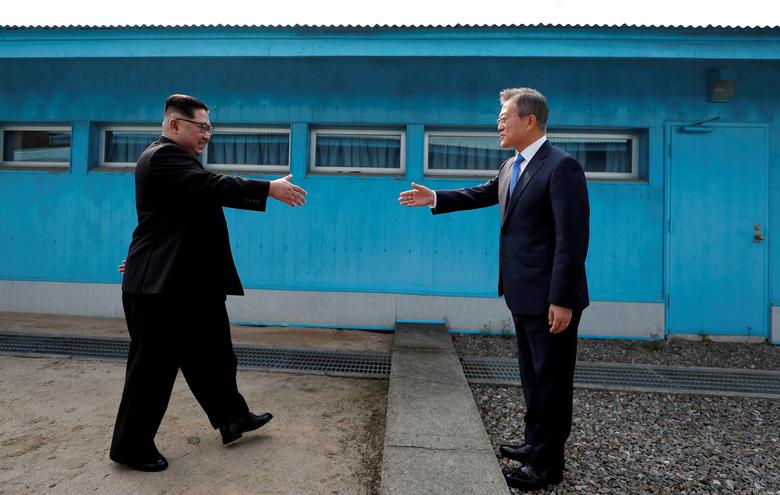}
  \caption{South Korean President Moon Jae-in and North Korean leader Kim Jong Un shake hands at the truce village of Panmunjom inside the demilitarized zone separating the two Koreas.}
\end{subfigure}%
\begin{subfigure}{.25\textwidth}
  \centering
  \captionsetup{width=.9\linewidth}
  \includegraphics[width=.9\linewidth]{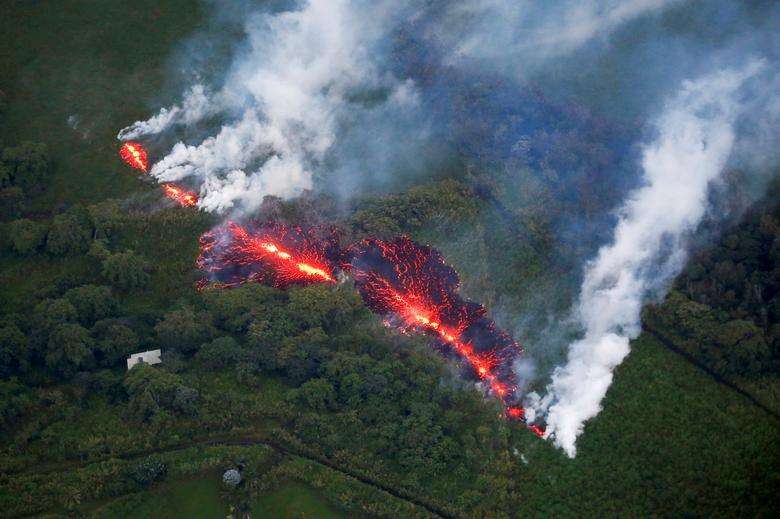}
  \caption{Lava erupts from a fissure east of the Leilani Estates subdivision during ongoing eruptions of the Kilauea Volcano.}
\end{subfigure}
\caption{Examples of image-claim pairs from The Reuters Dataset.}
\label{fig:reuters_examples}
\end{figure}

Yet, this lack of \emph{True}-labeled examples can pose some challenges for classification models and the evaluation process as well. This is why we decided to invest some time in gathering more \emph{True} examples as we explain below.

\subsection{The Reuters Dataset}
At the end of each year, Reuters publishes a list of about 100 photos, called \emph{Pictures of the Year}. Conveniently for us, each photo comes with a short textual description, which we can use as a \emph{claim}. We collected all of these pictures from four consecutive years: 2015, 2016, 2017, and 2018. As a result, we ended up with a total of 395 True image-claim pairs. Some examples are shown in Figure~\ref{fig:reuters_examples}. We further performed close manual inspection, and we did not find any obvious differences between these images compared to the ones from \emph{The Snopes Dataset}. In terms of the claim, texts from Reuters seem to be longer, but this should not be a problem, since we do not use the length as a feature.

\section{Experiments and Evaluation}
\label{sec:experiments}

\subsection{Setup}

Note that the above two datasets contain 1,233 examples combined, and these examples are relatively well-balanced: 592 \emph{True} and 641 \emph{False}. As this is a small size, we chose to test the performance of the models using cross-validation. If we mix the data from the two sources having in mind that the Reuters dataset has examples from the \emph{True} class only, we fear that the models might implicitly learn each example's source, not its factuality. Hence, we designed the following two cross-validation experiments:

\paragraph{Testing on Snopes-only data.} Ten times, using a different random seed, we do the following:
\begin{enumerate}
\item Randomly choose 50 \emph{True} and 50 \emph{False} Snopes examples and use them as a test set.
\item Use the rest of the \emph{True} Snopes data plus all Reuters data as \emph{True} training examples.
\item Randomly sample the necessary number of examples from the \emph{False} Snopes data, so that the training set is balanced.
\end{enumerate}

Finally, we compute the average of the evaluation measures for all ten folds.

\paragraph{Testing on Snopes + Reuters data.} Ten times, using a different random seed, we perform the following steps:
\begin{enumerate}
\item Combine all Snopes and Reuters data into a single dataset.
\item Balance the resulting dataset by randomly choosing the necessary number of \emph{False} examples.
\item Do a random train-test split, so that the test set contains 100 examples.
\end{enumerate}

As in the previous experiment, we compute the average of the evaluation measures for all ten folds.

\subsection{Classification model}

We used a Linear SVM with the default value of $C$=1. 
We trained a separate SVM model for each feature type, then we applied a softmax to normalize the values, and finally we averaged the confidences of the classifiers to make the final decision.

\subsection{Results}

We used the following evaluation measures:
\begin{itemize}
\item \textbf{Accuracy}, because the classes are balanced, and the majority-class baseline for all experiments is 50.0.
\item \textbf{Average Precision}, since it is useful if we want to have a ranking task, e.g.,~to prioritize which claims about images human fact-checkers should check first. Again, the random baseline for all experiments is 50.0.
\end{itemize}

\begin{table*}[tbh]
\centering
\begin{tabular}{lcccc}
  \toprule
  \bf Feature & \bf Acc (S) & \bf AP (S) & \bf Acc (S+R) & \bf AP (S+R)\\
  \midrule
All & \textbf{63.2} & \textbf{73.0} & \textbf{80.1} & \textbf{90.3} \\
\midrule
True media percentage & \textbf{62.1} & 59.3 & 74.6 & 69.8 \\
Embedding similarity of claim \& true media titles & 61.1 & 62.5 & 74.0 & 69.0 \\
Cosine similarity of claim \& true media titles & 61.1 & 58.4 & 73.8 & 69.0 \\
Known media percentage & 60.4 & 58.6 & 74.6 & 71.9 \\
URL domains & 60.3 &\textbf{ 67.3} & \textbf{78.6} & \textbf{89.7} \\
Google tags & 58.5 & 63.9 & 71.5 & 82.1 \\
Mixed media percentage & 58.0 & 56.1 & 62.4 & 60.8 \\
Claim text & 57.1 & 60.8 & 74.9 & 83.8 \\
True media titles & 55.8 & 63.1 & 73.6 & 81.4 \\
Mixed media titles & 55.4 & 58.5 & 63.1 & 67.2 \\
URL categories & 53.7 & 56.0 & 70.3 & 76.2 \\
False media titles & 50.3 & 50.8 & 50.6 & 50.4 \\
False media percentage & 49.9 & 51.1 & 50.4 & 50.4 \\
\midrule
Baseline & 50.0 & 50.0 & 50.0 & 50.0 \\
\bottomrule
\end{tabular}

\caption{\label{individualfeatures} Accuracy and Average Precision for individual feature types, calculated using 10-fold cross-validation using the Snopes dataset (\textbf{S}), and the Snopes+Reuters dataset (\textbf{S+R}).}
\end{table*}

Table~\ref{individualfeatures} illustrates the importance of each feature type in isolation. We can see that almost all individual feature types manage to outperform the two 50\% baselines. The only weak features are those related to false sources of information: percentage of unreliable media writing about the image and the words used in the titles of the articles. Moreover, using all features (with a model combination as explained above) works best: 63.2\% and 80.1\% Accuracy, 73.0\% and 90.3\% Average Precision for S and S+R, respectively. The top-3 feature types for the Snopes test set are \emph{true media percentage} (62.1\% for S and 74.6\% for S+R), \emph{embedding similarity} (61.1\% for S and 74.0\% for S+R), and \emph{cosine similarity} (61.1\% for S and 73.8\% for S+R). In either experiment, Average Precision is higher than Accuracy. Larger improvements are achieved for the Snopes + Reuters test set, which could be due to the model making more mistakes on the \emph{True} examples from Snopes and being better on \emph{True} examples from Reuters.

\begin{figure*}
\centering
\includegraphics[width=12cm,keepaspectratio]{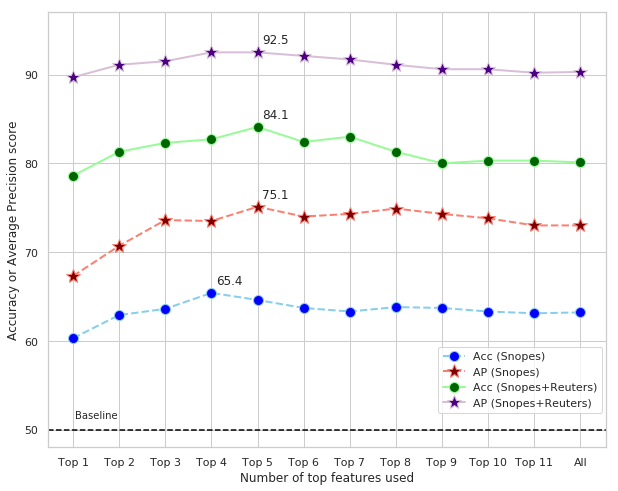}
\caption{\label{fig:plot-all}Accuracy and Average Precision on 10-fold cross-validation using top-$n$ features.}
\end{figure*}

Figure~\ref{fig:plot-all} shows combinations of the top-$n$ features using each feature's performance in terms of Average Precision. Note that these top features for the two experiments are different: we use the scores in the AP(S) column in Table~\ref{individualfeatures} for the Snopes dataset, and the AP(S+R) column for the Snopes+Reuters dataset. We can see that selecting the top 4 to 5 features works best, yielding 65.4\%, 75.1\%, 84.1\% and 92.5\%. 

Note that the Average Precision scores are higher than those for Accuracy, and the scores for the Snopes+Reuters dataset are higher.

\section{Discussion}
\label{sec:discuss}

\begin{figure}
\centering
\includegraphics[width=8cm,keepaspectratio]{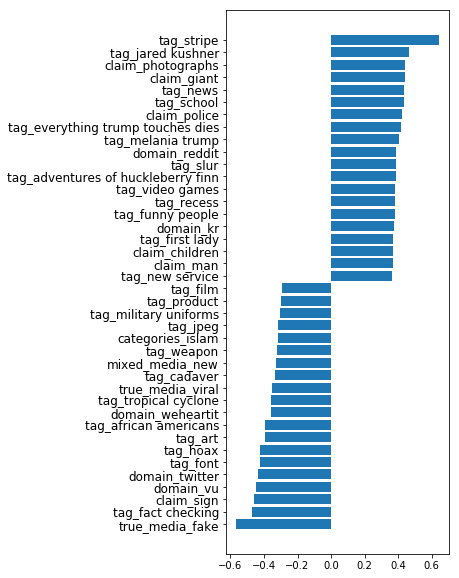}
\caption{\label{fig:feature-importance} The most informative features: 20 positive and 20 negative. Prefixes indicate feature types.}
\end{figure}

\subsection{Most Important Individual Features}
Above, we explored the performance of individual feature groups. Here we try to understand what the most important individual features are. For this purpose, we trained a model on all features, and then we analyzed the weight of each feature in this full model. Note that this is different from the setup in the previous section, where we trained a separate model for each feature group, and then we combined the predictions of these models in an ensemble; in contrast, here we just put all features from all groups together.
The results are visualized in Figure~\ref{fig:feature-importance}. We can see that some of them seem random, e.g.,~\textit{adventures of huckleberry fin} or \textit{everything trump touches dies}. However, there are a few that signal false information, e.g.,~words like \textit{fake} and \textit{viral} mentioned in the title of a trustworthy medium, or tags like \textit{hoax} and \textit{fact-checking}. The existence of images in the dataset that were modified for artistic purposes can explain tags such as \textit{art} and \textit{film}. Also, according to our best features, we should not trust much images that appear on \textit{Twitter} or ones related to sensitive topics like \textit{african americans} or \textit{islam}.

\subsection{What Did Not Work}

{\bf Metadata from images}:
In an attempt to capture possible manipulation of the input image, we gathered meta information using an open-source tool\footnote{\url{http://github.com/redaelli/imago-forensics}} for image forensics. The tool extracts metadata in the form of about 100 features such as size, resolution, GPS location. However, most of this metadata turns out to be missing from our images: only five features could be extracted for more than half of the images from the Snopes dataset.

\begin{figure}
\centering
\includegraphics[width=8cm,keepaspectratio]{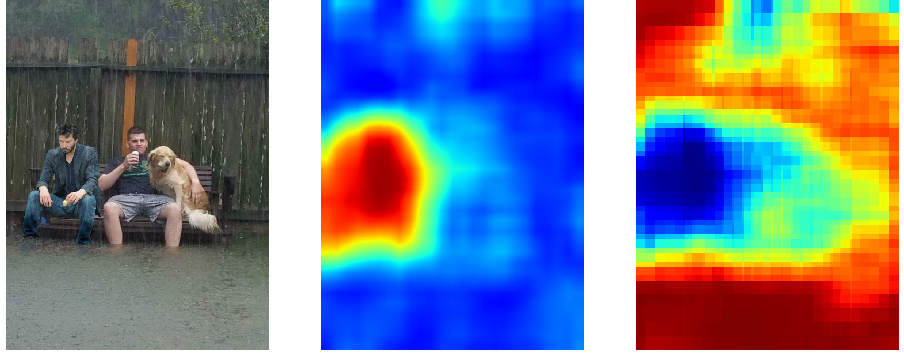}
\caption{\label{fig:self-cons-keanu}Predicted clusters for one of the images in the Self-Consistency paper. Keanu Reeves has been spliced into the photo and his body was separated correctly by both MeanShift and DBSCAN.}
\end{figure}

{\bf Image Splice Detection}:
As we have already mentioned, one of the reasons why an image could be fake is that it has been digitally manipulated. A common manipulation is splicing, i.e.,~cropping and stitching together parts of the same image or multiple different images.
We explored an approach that looks for the lack of self-consistency in images and outputs clusters of the predicted image parts using two algorithms: MeanShift and DBSCAN \cite{10.1007/978-3-030-01252-6_7}. An illustration on how it works is shown in Figure~\ref{fig:self-cons-keanu}. We decided to validate the method by using a pretrained model,\footnote{\url{http://github.com/minyoungg/selfconsistency}} which we applied to some images from the Snopes dataset that were obvious cases of splicing. 

\begin{figure}
\centering
\includegraphics[width=8cm,keepaspectratio]{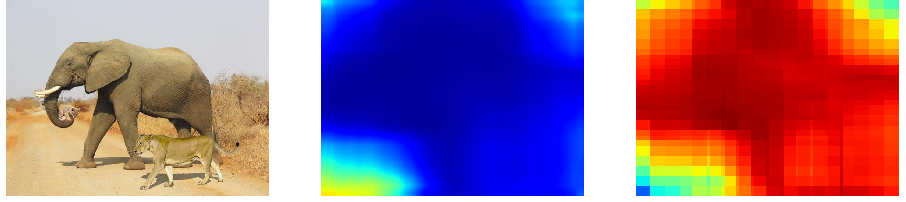}
\caption{\label{fig:self-cons-lion}Predicted clusters for one of the images in the Snopes Dataset. The original image depicts an elephant; the lion and the cub have been photoshopped on top (\href{http://www.snopes.com/fact-check/elephant-carrying-lion-cub}{Source}). The clustering algorithms did not detect this splicing.}
\end{figure}

Unfortunately, this seemed not to work for us. Figure~\ref{fig:self-cons-lion} shows an example where the model could not find the spliced regions.
Eventually, we abandoned this direction as the inference time and the required resources were significant, and the performance was not very good on our dataset.

{\bf Error Level Analysis}:
Error Level Analysis (ELA) helps to identify areas within an image that are at different compression levels. With JPEG images such as the ones in our Snopes and Reuters datasets, the entire image should be at roughly the same level. If a section of the image is at a significantly different error level, this would indicate a likely digital modification.

\begin{figure}
\centering
\includegraphics[width=7cm,keepaspectratio]{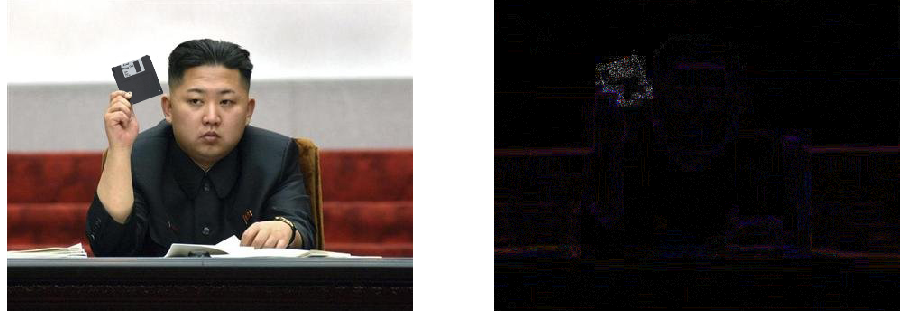}
\caption{\label{fig:ela-floppy}One of the ELA examples on \url{http://fotoforensics.com}. The part of the image with a floppy disk appears brighter on the ELA map, as it was spliced on top of the original.}
\end{figure}

\begin{figure}
\centering
\includegraphics[width=7cm,keepaspectratio]{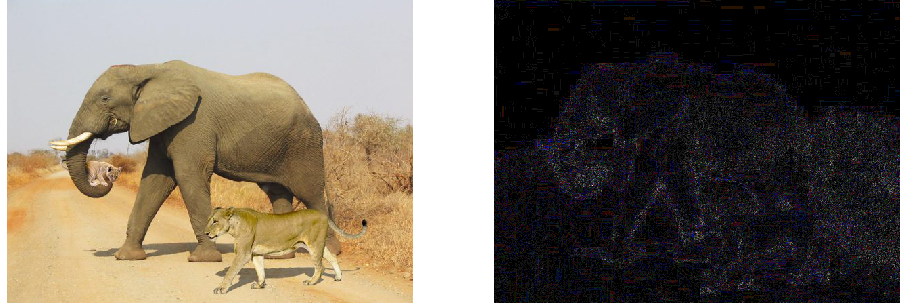}
\caption{\label{fig:ela-lion} ELA applied to one of the images from the Snopes dataset. The spliced regions, i.e.,~the lion and the cub, could not be identified.}
\end{figure}

ELA works by intentionally resaving the image at a known error rate such as 95\%, and then computing the difference between the images. If there is virtually no change, then the cell has reached its local minima for error at that quality level. However, if there is a large change, then the pixels are not at their local minima and are effectively original. This method can be used to identify splicing, because stitched regions will appear brighter on the ELA version of the image. This is illustrated in Figure~\ref{fig:ela-floppy}. After manual inspection of ELA versions of images from our dataset, we did not find the method to be very promising, see Figure~\ref{fig:ela-lion}.
% yet, we decided to try it anyway. 
% We fed the ELA images into a one-layer Convolutional Neural Net and we got 54.4\% average accuracy on the Snopes-only Test Set and 57.5\% average accuracy on the Snopes+Reuters dataset.\todo{Well, then this actually works! It does beat the baseline! Why not add it to the mix? This would make the approach more multi-modal. Why do we say this did not work?? BTW, how about the other methods that ``did not work''?}

\subsection{Testing on New Data}
All of the experiments described so far were performed on claim-image pairs from Snopes that were published in the period between November 20, 2000 and February 1, 2019. The data from February up until April 29, 2019 has been left untouched, which makes it suitable for performing one final test of the developed system. In these three months, 64 articles were published in the Fauxtography section, of which 14 were labeled as \emph{True} and 25 as \emph{False}. To balance this new test set, we subsampled 14 \emph{False} examples randomly. The training was performed on all previously collected data from Snopes and Reuters, balanced in the same way. For better certainty of the performance, we sampled randomly the training and the test sets ten times, and we report the average scores. 

The results when using the top features based on the Average Precision for the Snopes dataset are shown in Table \ref{testresults}. We can see that the best Average Precision is achieved by using the single top feature of URL domains: 71.7\%. When we add to this the second best one, i.e.,~the Google tags, we get an Accuracy of 64.3\%. The scores of the models that use more than three features are not displayed since they were not as good.

The best-performing features across the experiments differ, but as Table \ref{individualfeatures} shows, the \emph{URL domains} are top-1 in three out of four experiments, and \emph{claim text} is top-2 in two out of four experiments.

\begin{table}[tbh]
\begin{center}
\begin{tabular}{lcc}
  \toprule
  \bf Features & \bf Acc & \bf AP \\
  \midrule
All & 59.3 & 69.7 \\
\midrule
Top 1 & 62.9 & \textbf{71.7} \\
Top 2 & \textbf{64.3} & 70.4 \\
Top 3 & 57.5 & 70.6  \\
\midrule
Baseline & 50.0 & 50.0\\
\bottomrule
\end{tabular}
\end{center}
\caption{\label{testresults} Accuracy and Average Precision on the New test dataset.}
\end{table}

\section{Conclusion and Future Work}
\label{sec:future}

We have presented our efforts towards fighting Fauxtography, namely detecting fake claims about images, which is an under-explored research direction. In particular, we created a new dataset for this problem, and we explored a variety of features modeling the claim, the image, and the relationship between the two. The evaluation results have shown sizable improvements over the baseline. We release our dataset,\footnote{\url{http://gitlab.com/didizlatkova/fake-image-detection}} hoping to enable further research on fact-checking claims about images.

In future work, we plan to extend the dataset with more examples, to try other features, e.g.,~from social media and from metadata,\footnote{The lack of metadata that we observed can be explained by the fact that Snopes.com is not the original source of the image files; it collected images from various external sources. Those sources might not be the original creator either and multiple downloading and uploading of files, with possible reformatting could mean loss of metadata as many Web sites reformat images and/or delete/change the metadata of the images uploaded to it. Finally, we could not extract any EXIF metadata for the Reuters images, even though we got them from Reuters. Yet, maybe the metadata can be recovered using Reverse Image Search.}  and to adapt the system to work with other languages. We further plan  experiments with fact-checking claims about videos.

\section*{Acknowledgements}

This research is part of the Tanbih project,\footnote{\url{http://tanbih.qcri.org/}} which aims to limit the effect of ``fake news'', propaganda and media bias by making users aware of what they are reading. The project is developed in collaboration between the Qatar Computing Research Institute (QCRI), HBKU and the MIT Computer Science and Artificial Intelligence Laboratory (CSAIL).

\bibliographystyle{acl_natbib}
\bibliography{emnlp-ijcnlp-2019}

\appendix

\end{document}